\documentclass[
]{ceurart}

\usepackage{url}
\usepackage{subcaption}
\usepackage{float}
\usepackage{graphicx}
\usepackage{subcaption}
\usepackage{geometry}
\usepackage{array}
\usepackage{longtable}
\usepackage{soul}
\usepackage{tablefootnote} 
\floatstyle{ruled}
\newfloat{prompt}{thp}{lop}
\floatname{prompt}{Prompt}

\usepackage{color}

\usepackage{listings}
\begin{document}

\copyrightyear{2024}
\copyrightclause{Copyright for this paper by its authors.
  Use permitted under Creative Commons License Attribution 4.0
  International (CC BY 4.0).}

\conference{CLEF 2024: Conference and Labs of the Evaluation Forum, September 09–12, 2024, Grenoble, France}

\title{FactFinders at CheckThat! 2024: Refining Check-worthy Statement Detection with LLMs through Data Pruning}

\title[mode=sub]{Notebook for the CheckThat! Lab at CLEF 2024}

\author[1]{Yufeng Li}[%
orcid=0009-0008-8740-4994,
email=yufeng.li@qmul.ac.uk
]
\cormark[1]
\fnmark[1]
\address[1]{School of Electronic Engineering and Computer Science, Queen Mary University of London}

\author[1]{Rrubaa Panchendrarajan}[%
orcid=0000-0002-1403-2236,
email=r.panchendrarajan@qmul.ac.uk
]
\cormark[1]
\fnmark[1]

\author[1]{Arkaitz Zubiaga}[%
orcid=0000-0003-4583-3623,
email=a.zubiaga@qmul.ac.uk,
url=www.zubiaga.org,
]

\cortext[1]{Corresponding author.}
\fntext[1]{These authors contributed equally.}

\begin{abstract}
The rapid dissemination of information through social media and the Internet has posed a significant challenge for fact-checking, among others in identifying check-worthy claims that fact-checkers should pay attention to, i.e. filtering claims needing fact-checking from a large pool of sentences. This challenge has stressed the need to focus on determining the priority of claims, specifically which claims are worth to be fact-checked. Despite advancements in this area in recent years, the application of large language models (LLMs), such as GPT, has only recently drawn attention in studies. However, many open-source LLMs remain underexplored. Therefore, this study investigates the application of eight prominent open-source LLMs with fine-tuning and prompt engineering to identify check-worthy statements from political transcriptions. Further, we propose a two-step data pruning approach to automatically identify high-quality training data instances for effective learning. The efficiency of our approach is demonstrated through evaluations on the English language dataset as part of the check-worthiness estimation task of CheckThat! 2024. Further, the experiments conducted with data pruning demonstrate that competitive performance can be achieved with only about 44\% of the training data. Our team ranked first in the check-worthiness estimation task in the English language.
\end{abstract}


\maketitle

\section{Introduction}
With the significant development of the Internet and social media over the past decades, the practical challenges associated with fact-checking have become more complex \cite{zeng2021automated,guo2022survey}. Social media platforms have facilitated the rapid dissemination of information, which increases the difficulty of distinguishing misinformation from accurate information \cite{allcott2019trends}. Concurrently, the general agreement on what should be fact-checked has expanded to include online content and claims made by politicians, resulting in a wide range of claims to be verified. As the initial step in the fact-checking process, claim detection plays a crucial role in efficiently identifying check-worthy claims, allowing for quicker progression to subsequent stages of verification \cite{panchendrarajan2024claim}. Therefore, research on check-worthy claim detection is essential for advancing the field of fact-checking, where the CheckThat! shared task has played a significant role in recent years \cite{nakov2018overview}.

Since the beginning of the CheckThat! competition, traditional machine learning models and neural network models have been commonly employed for the task of claim check-worthiness detection. At CheckThat! 2018, the top submission was achieved by a team using Support Vector Machines and Multilayer Perceptrons \cite{atanasova2018overview}, while the highest scores at CheckThat! 2019 were obtained using Long Short-Term Memory (LSTM) networks \cite{atanasova2019overview}. Although BERT \cite{vaswani2017attention} was introduced in 2018, the exploration of this transformer-based model for claim check-worthiness detection began only in 2020. In that year, the team utilizing RoBERTa secured the first position in the English category \cite{shaar2020overview}.

Beyond the application of machine learning models, various techniques have been explored throughout the CheckThat! competition. Feature representation methods, including word embeddings, Bag of Words, Named Entity Recognition, and Part of Speech tagging \cite{atanasova2018overview, atanasova2019overview, shaar2020overview} have been widely used to enhance model understanding of the task. More sophisticated representation techniques, such as LIWC \cite{shaar2021overview} and ELMo \cite{nakov2022overview}, have also been investigated. Additionally, statistics related to word usage, such as subjectivity and sentiment, have been incorporated. To address the challenge of imbalanced datasets, data augmentation strategies have been explored, with common methods including machine translation and sampling \cite{panchendrarajan2024claim}.

Large language models (LLMs) have seen remarkable advancements in recent years, with GPT \cite{brown2020language} models predominantly utilized as the latest and most effective solution in CheckThat! competitions. Several teams have shown competitive and winning performance of the model in various CheckThat! tasks, including check-worthiness estimation in multiple languages \cite{agresti2022polimi, sawinski2023openfact}. Although GPT models have demonstrated competitive performance in CheckThat! tasks, their fine-tuning and inference entail associated costs. Simultaneously, numerous open-source LLMs have also demonstrated substantial advancements showing equivalent performance to GPT models. This enables the global community of researchers to benefit by transferring the knowledge of these powerful models cost-effectively by fine-tuning them on various downstream tasks. While fine-tuned BERT-based and GPT models have been extensively and routinely examined in the domain of check-worthiness estimation \cite{panchendrarajan2024claim}, open-source LLMs, as emerging language models, have not yet been thoroughly investigated within this specific field. Therefore, this study aims to explore a wide range of open-source LLMs while leveraging their capabilities through prompt engineering for check-worthiness estimation.  

This paper presents the experiments conducted for CheckThat! 2024 task 1 \cite{clef-checkthat:2024-lncs,clef-checkthat:2024:task1}, check-worthiness estimation in the English language. The task involves identifying check-worthy statements from political transcriptions. Drawing inspiration from the impressive performance of  LLMs in the recent CheckThat! competitions, we explore eight popular open-source LLMs, specifically Llama2-7b, Llama2-13b \cite{touvron2023llama}, Llama3-8b, Mistral \cite{jiang2023mistral}, Mixtral \cite{jiang2024mixtral}, Phi3-Mini-4K \cite{abdin2024phi}, Falcon \cite{almazrouei2023falcon}, and Gemma-7b \cite{team2024gemma} with prompt engineering for identifying check-worthy statements. Considering the noisy and imbalanced nature of the training data, we propose a two-step data pruning process to isolate high-quality training data instances for effective learning with LLMs. Especially, we identify the informative sentences first and apply an under-sampling technique, Condensed Nearest Neighbour \cite{hart1968condensed}, to create a balanced training dataset. Our fine-tuned Llama2-7b \cite{touvron2023llama} model on the original training data shared by the task organizers scored the highest F1-score in the task 1 leaderboard in the English language. However, the experimental results indicate that similar or better performance can be achieved with data pruning techniques while retaining only about 44\% of high-quality data instances from the original training data. Furthermore, this approach resulted in a reduction in fine-tuning time by a similar proportion, which could significantly lower the resource demands for fine-tuning larger models. All relevant source code and data are available on GitHub,\footnote{\url{https://github.com/isyufeng/FactFinders}} and the fine-tuned model can be accessed on Huggingface.\footnote{\url{https://huggingface.co/Rrubaa/factFinders-checkworthy-estimation}}

The remainder of the paper is structured as follows. Section \ref{sec:methodology} presents the methodology with the introduction to the LLMs experimented, prompts used and the data pruning techniques proposed. Section \ref{sec:Results} discusses the experiment results, followed by section \ref{sec:conclusion} concluding the key findings and future directions.

\section{Methodology}\label{sec:methodology}
Our goal was to automatically refine the training data to obtain high-quality training data instances and fine-tune open-source Large Language Models (LLMs) to identify check-worthy statements from political transcriptions. This section introduces the dataset, LLMs used in the experiments, prompt engineering carried out, fine-tuning process, and the two-step data pruning we applied to the training data for effective learning.   

\subsection{Dataset}\label{sec:dataset}
\begin{figure}[]
    \centering
    \includegraphics[width=0.9\linewidth]{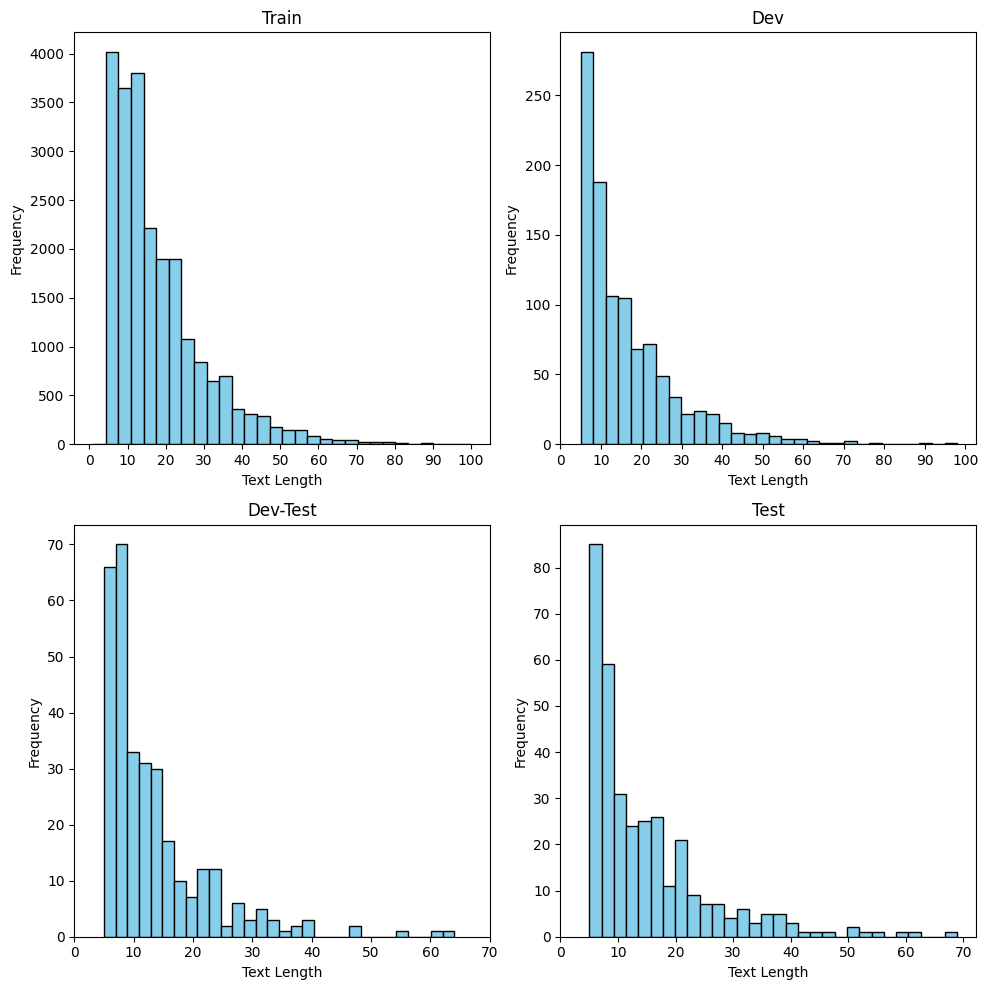}
    \caption{Distribution of Text Length in each Partition}
    \label{fig:data-distribution}
\end{figure}

The dataset provided by CheckThat! 2024 comprises the \textit{train}, \textit{dev}, and \textit{dev-test} partitions, containing 23,849 sentences from political transcriptions, along with a later release of \textit{test} partition, bringing the total to 24,163 sentences. Table \ref{tab:data_statisitcs} presents the statistics for each partition. From this table, it is evident that the dataset is imbalanced, posing a challenge for the check-worthy statement detection task. Furthermore, Figure \ref{fig:data-distribution} illustrates the distribution of text lengths across each partition, revealing that the median length of the sentences is approximately 10-14 words in each partition, with 28\%-42\% of sentences containing fewer than 10 words. This indicates that the dataset not only suffers from class imbalance but also contains predominantly short sentences, hence implying a limited amount of information.

\begin{table}[]
\caption{Statistics of the Dataset}
\begin{tabular}{llll}
\toprule
Partition         & Check-worthy & Non-check-worthy & Total \\ \midrule
Train    & 5,413         & 17,086            & 22,499 \\
Dev      & 238          & 794              & 1,032     \\
Dev-Test & 108          & 210              &  318     \\
Test     & 88           & 253              &  341   \\
\bottomrule
\label{tab:data_statisitcs}
\end{tabular}
\end{table}

\subsection{LLMs for Check-worthy Statement Detection}
Open-source LLMs offer substantial advantages in terms of cost, transparency, community support, and ethical considerations. Consequently, we investigated eight prominent open-source LLMs, as detailed in Table \ref{tab:model_info} by fine-tuning them for check-worthy statement detection.

\subsubsection{Large Language Models}
Llama2-7b, Llama2-13b \cite{touvron2023llama}, and Llama3-8b are part of the Llama family, developed by Meta, and are available in various sizes. The most recent release, Llama3-8b, was made available in April 2024. These models have been optimized for text generation and dialogue applications. Similarly, Mistral \cite{jiang2023mistral} and Mixtral \cite{jiang2024mixtral} are both developed by Mistral AI. Mistral mainly focuses on optimizing transformer models for language tasks, achieving high efficiency and performance in a compact form. Mixtral, with its hybrid approach, aims to integrate the best of various AI methodologies, offering flexibility and scalability for complex applications. Compared to this bigger model, Phi3-Mini-4K \cite{abdin2024phi} is a smaller variant of the Phi-3-mini, developed by Microsoft, designed to provide capabilities similar to its larger counterparts but with a reduced number of parameters, making it more accessible and easier to run on less powerful hardware. Similarly, Falcon \cite{almazrouei2023falcon}, developed by TII, stands as one of the most powerful open-source models and consistently achieves top positions on the OpenLLM leaderboard hosted on Hugging Face. One of the latest models we experimented with, Gemma-7b \cite{team2024gemma} belongs to the Gemma family developed by Google DeepMind, which is designed to offer a balance between computational efficiency and advanced capabilities in generating text and understanding complex language queries. We fine-tuned these eight open-source LLMs published in Huggingface platforms (links listed in Table \ref{tab:model_info}) for check-worthy statement detection from political transcriptions.  

\subsubsection{Prompt Engineering}\label{sec:prompt-engeering}

\begin{table}[t]
\centering
\caption{Open-Source LLMs Used.}
\begin{tabular}{lll}
\toprule
Model & Number of Parameters & Release Date \\ \midrule
Llama2-7b\tablefootnote{\url{https://huggingface.co/meta-llama/Llama-2-7b-hf}} & 7 billion & July 2023 \\ 
Llama2-13b\tablefootnote{\url{https://huggingface.co/meta-llama/Llama-2-13b-hf}} & 13 billion & July 2023 \\ 
Llama3-8b\tablefootnote{\url{https://huggingface.co/meta-llama/Meta-Llama-3-8B-Instruct}} & 8 billion & April 2024 \\ 
Mistral\tablefootnote{\url{https://huggingface.co/mistralai/Mistral-7B-Instruct-v0.2}} & 7 billion & September 2023 \\ 
Mixtral\tablefootnote{\url{https://huggingface.co/mistralai/Mixtral-8x7B-v0.1}} &  45 billion & December 2023 \\ 
Phi-3-Mini-4K\tablefootnote{\url{https://huggingface.co/microsoft/Phi-3-mini-4k-instruct}} & 3.8 billion & April 2024 \\ 
Falcon\tablefootnote{\url{https://huggingface.co/tiiuae/falcon-7b}} & 7 billion & March 2023 \\
Gemma-7b\tablefootnote{\url{https://huggingface.co/google/gemma-7b}} & 7 billion & February 2024 \\
\bottomrule
\label{tab:model_info}
\end{tabular}
\end{table}

Given the critical role of prompts in the performance of LLMs, we initially came up with a simple yet direct prompt, as illustrated in Prompt \ref{prompt-v2}. Observing that this initial prompt resulted in lengthy responses with redundant information in the zero-shot setting, and lacked a clear definition of check-worthiness, we employed ChatGPT-4 to refine and improve the prompt, resulting in Prompt \ref{prompt-v1}. All eight LLMs were fine-tuned using this refined prompt to generate `Yes' or `No' answers indicating the check-worthiness of the input statement. 
Furthermore, we observed that a significant proportion of sentences in the training data utilized pronouns to refer to political entities, thereby increasing uncertainties and ambiguities. Therefore, we experimented with an expanded version of Prompt \ref{prompt-v1} (i.e. Prompt \ref{prompt-v3}) to evaluate whether explicitly indicating that the pronouns in the input statement may refer to political entities could enhance the performance of the fine-tuned model. However, the initial experiments on prompt engineering revealed that neither the compressed prompt (Prompt \ref{prompt-v2}) nor the expanded (Prompt \ref{prompt-v3}) improved the performance of fine-tuned models (refer to Table \ref{tab:performance_prompt}). 

\subsubsection{Effective Fine-tuning}
Fine-tuning an LLM is a challenging task due to the resource requirements, especially the memory demand. While this challenge can be escalated by training only certain layers of the LLM, still the computational requirement associated with gradient updates requires a lot of GPU memory. Therefore we use the Low-Rank Adaption technique (LoRA) for fine-tuning the LLMs. Instead of updating the weights directly, LoRA keeps track of the changes through low-rank perturbations requiring only minimal GPU memory. The LoRA configuration used for the fine-tuning is listed in Section \ref{sec:hyper-paramters}. To ensure experimental control, consistent hyperparameters were applied across all eight LLMs (see Table \ref{tab:parameters}). 

\vspace{5mm}
\noindent The performance of each model on check-worthy statement detection is presented in Table \ref{tab:llm_comparison}. Unfortunately, we could only compare the performance of the Llama family, Mistral, and Mixtral models during the testing phase of the competition. Therefore, the Phi-3-Mini-4k, the best-performing model in the Dev-Test partition wasn't considered for the remaining experiments. Considering the competitive performance of the Llama2-7b model and the time and memory required, the rest of the experiments were carried out by fine-tuning the Llama2-7b using Prompt \ref{prompt-v1}.  

\begin{prompt}[t]
\begin{small}
\begin{verbatim}
### Instruction:
Read the statement provided as below.
Your task is to evaluate whether the statement contains information or claims 
that are worthy to be verified through fact-checking. If the statement presents 
assertions, facts, or claims that would benefit from verification, respond with 
'Yes'. If the statement is purely opinion-based, trivial, or does'not contain 
any verifiable information or claims, respond with 'No'.

### Input Sentence: <input sentence>

### Response: <Yes/No>
\end{verbatim}
  \caption{Check-worthy Statement Detection}
  \label{prompt-v1}
\end{small}
\end{prompt}


\begin{prompt}[t]
\begin{small}
\begin{verbatim}
### Instruction:
Classify the following statement as check-worthy (Yes) or not check-worthy (No).

### Input Sentence: <input sentence>

### Response: <Yes/No>
\end{verbatim}
  \caption{Check-worthy Statement Detection - Compressed}
  \label{prompt-v2}
\end{small}
\end{prompt}

\begin{prompt}[t]
\begin{small}
\begin{verbatim}
### Instruction:
Read the statement provided below from political transcriptions.
The pronouns in the statement may refer to political entities. 
Your task is to evaluate whether the statement contains information or claims 
that are worthy to be verified through fact-checking. If the statement presents 
assertions, facts, or claims that would benefit from verification, respond with 
'Yes'. If the statement is purely opinion-based, trivial, or does'not contain 
any verifiable information or claims, respond with 'No'.

### Input Sentence: <input sentence>

### Response: <Yes/No>
\end{verbatim}
  \caption{Check-worthy Statement Detection - Expanded}
  \label{prompt-v3}
\end{small}
\end{prompt}

\subsection{Data Pruning for Effective Learning}
As previously mentioned, the training data is highly imbalanced,  mostly containing shorter sentences with limited information. Recent studies \citep{sawinski2023openfact} have demonstrated that instead of using the entire training data for fine-tuning, using only the high-quality labels improves the performance of check-worthy statement detection from political transcriptions. Inspired by this direction, we experimented with a two-step data pruning approach to automatically identify high-quality data instances for effective learning.

\subsubsection{Step 1 - Identifying Informative Sentences}
We began with identifying informative sentences in the training data that could potentially convey meaningful information for the training. In other words, we intended to remove the noisy instances from the training data as the first step of the data-pruning process. We define a political statement as informative if it meets one of the following four criteria:
\begin{itemize}
    \item \textit{Check-worthy status is "Yes"}: If the class label of the statement is "Yes", then it is informative.
    \item \textit{Contains a named entity}: If the statement contains a named entity, it is highly likely to discuss information related to that entity. Hence the statement is informative.
    \item \textit{Contains an informative verb}: If the statement contains an informative verb, it is highly likely to discuss an informative action. Hence the statement is informative. 
    \item \textit{Lengthy enough}: If the statement is lengthy enough, it is likely to convey meaningful information. 
\end{itemize}

While the first criteria \textit{check-worthy status} is straightforward, the other criteria require extracting further information to determine the informative status of a sentence. We used the BERT model \cite{bert} fine-tuned\footnote{\url{https://huggingface.co/dslim/bert-base-NER}} for the Named Entity Recognition (NER) task to identify the presence of a named entity. The NER model identifies four types of named entities, person, organization, location, and miscellaneous from the input text. We noticed that only $37.8\%$ of the training data contains at least one named entity and only  $12.2\%$ of the training data mentions a person's name. This indicates the prevalent use of pronouns to refer to the political entities in the transcriptions, which makes the task more challenging.

In order to identify informative verbs, we first extracted all the verbs presented in the training data using the Part-of-Speech tagger from NLTK library.\footnote{\url{https://www.nltk.org/api/nltk.tag.pos_tag.html}} Extracted verbs were lemmatized further to bring them to their base form. This resulted in 3838 verbs in the training data to be identified as informative or not. We wanted to automatically classify each verb as either informative or not based on whether it conveys any check-worthy action. However, performing this binary classification in a zero-shot setting using a language model is a challenging task as explicitly defining an \textit{informative} verb in the prompt may result in ambiguous classification. Therefore, we performed a fine-grained categorization of the verbs into the following 10 categories. 
\begin{enumerate}
    \item \textit{Physical Actions}: e.g. Run
    \item \textit{Mental Actions}: e.g. Think
    \item \textit{Changes in State}: e.g. Grow
    \item \textit{Creation or Destruction}: e.g. Build
    \item \textit{Communication}: e.g. Discuss
    \item \textit{Movement}: e.g. Walk
    \item \textit{Emotion}: e.g. Hope
    \item \textit{Perception}: e.g. See
    \item \textit{Linking verbs }: e.g. is, has
    \item \textit{None}: Any verb that does not fit into the other categories
\end{enumerate}

The first 8 categories were obtained by prompting ChatGPT with the question "What are the types of action verbs?". In addition to the 8 action verb categories, we added \textit{Linking verbs} and \textit{None} resulting in 10 categories of verbs. The option "None" was added to the categories to indicate that the verb does not fit into any of the other 9 categories. 

\begin{prompt}[t]
\begin{small}
\begin{verbatim}
Classify the verb '<verb>' into one of the following types:
1. Physical Action
2. Mental Action
...
..
10. None

Verb type:
\end{verbatim}
  \caption{Verb Classification}
  \label{prompt-verb-classification}
\end{small}
\end{prompt}

We utilized Mixtral \cite{jiang2024mixtral} in a zero-shot setting to classify a verb into one of the 10 categories using the Prompt \ref{prompt-verb-classification}. Among the 10 categories, we chose the verb types \textit{Physical Actions}, \textit{Changes in State}, \textit{Creation or Destruction}, \textit{Communication}, and \textit{Movement} as the informative verbs, as the other verb types are less likely to represent a check-worthy action.      

\begin{figure}
  \centering
  \includegraphics[width=0.8\linewidth]{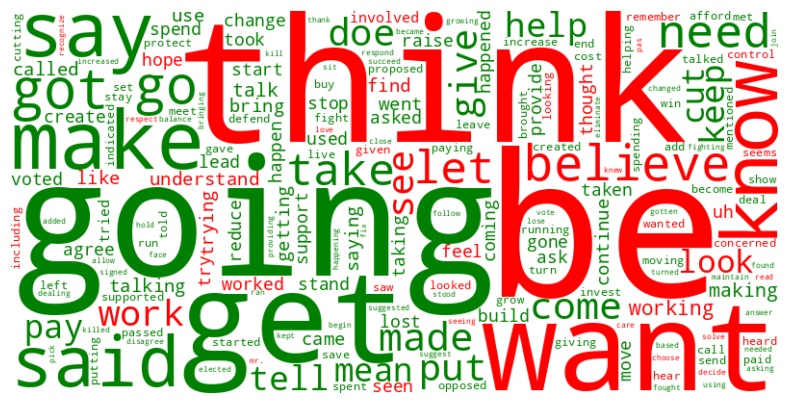}
  \caption{Word cloud indicating verb types and their frequencies in the training data.}
  \label{verb_type_cloud}
  
\end{figure}

\begin{figure}
  \centering
  \includegraphics[width=0.7\linewidth]{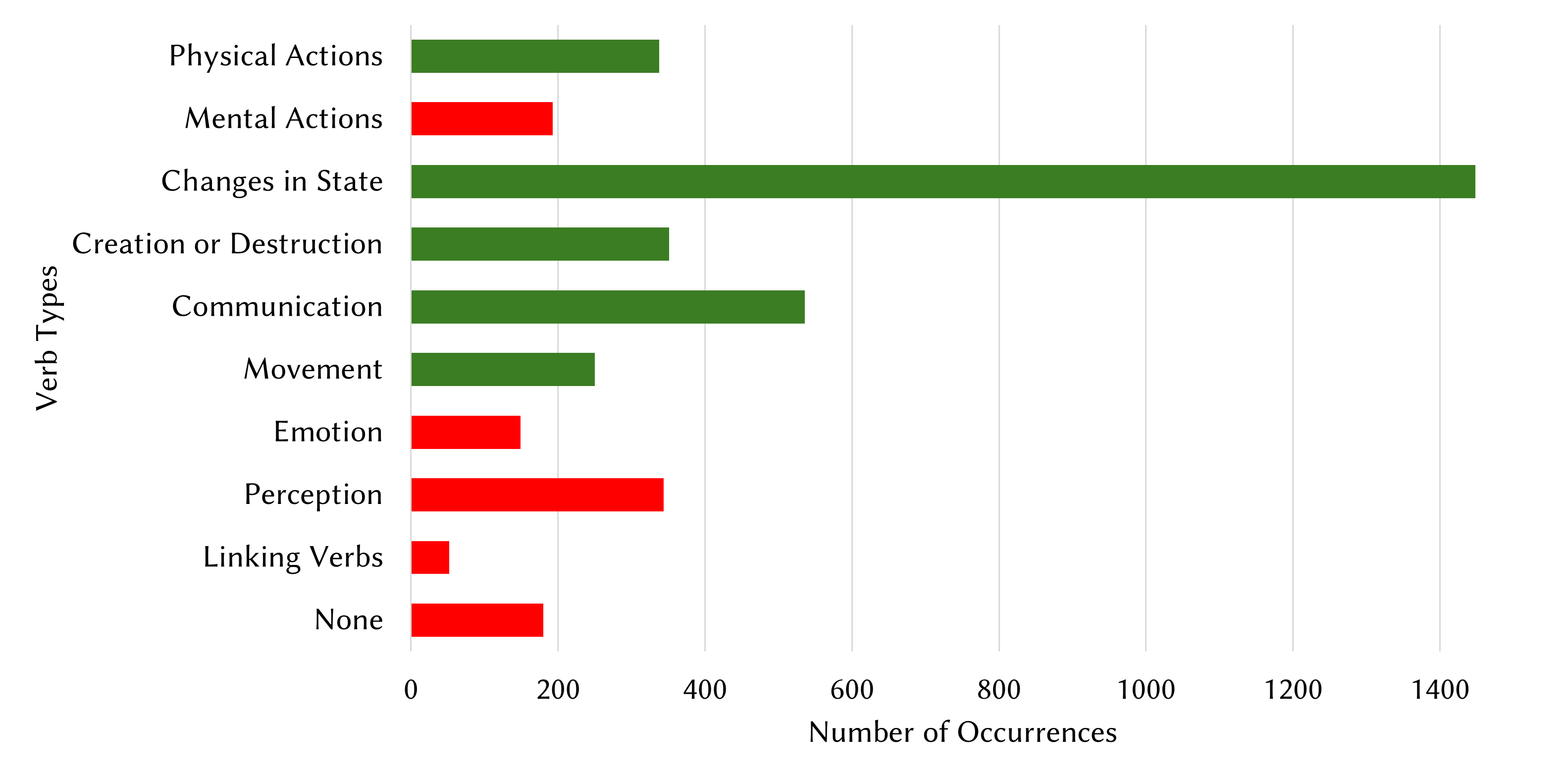}
  \caption{Distribution of verb types in the training data}
  \label{verb_type_distribution}
\end{figure}

Figure \ref{verb_type_cloud} shows the word cloud of verbs present in the training data. Informative verbs are highlighted in green color and non-informative verbs are highlighted in red color in the Figure. It can be observed that most of the verbs are classified into the two groups correctly. Figure \ref{verb_type_distribution} presents the verb type distribution in the training data. Interestingly the occurrences of informative verb categories are relatively high compared to non-informative verb categories. We further noticed that $88.2\%$ of the check-worthy statements contain at least one informative verb in the training data, whereas this value drops to $77.2\%$ for non-check-worthy statements. 

The final factor determining an informative sentence, \textit{minimum length} is difficult to define explicitly. Therefore, we choose the \textit{minimum length} value that reaches the optimal F1-score in the dev-test data by varying the value from 3 - 10. We excluded the stop words\footnote{\url{https://www.nltk.org/search.html?q=stopwords}} while calculating the length of a statement. We observed that the most informative sentences are obtained when the \textit{minimum length} factor is set to 8 (refer to Section \ref{sec:results_data_pruning} for the results). With this optimal setting, the first step of data pruning resulted in a reduced training data with 20,141 sentences. In other words, 2358 sentences ($10.5\%$ of original training data) were filtered out as non-informative sentences at this stage of the data pruning process. 

\subsubsection{Step 2 - Under Sampling using Condensed Nearest Neighbour}
The informative sentences identified in the previous stage are still imbalanced in class. Therefore, we executed an under-sampling technique, Condensed Nearest Neighbour (CNN) \cite{hart1968condensed}, to generate class-balanced training data. We retained all the minority data instances (check-worthy statements) and sampled only the majority data instances (non-check-worthy statements). The idea behind CNN sampling is to identify a subset of data instances that can be used to correctly classify all the other unsampled data instances using the 1-Nearest Neighbour rule. This makes sure that the sampled data distribution is the same as the original data distribution without any information loss. 

CNN requires the input data to be represented as vectors to iteratively sample data points from a vector space and perform the 1-Nearest Neighbour classification in the unsampled data points. We used BERT \cite{bert} embeddings to convert the sentences in the training data to a vector representation of length 768. The \textit{imbalanced}\footnote{\url{https://imbalanced-learn.org/stable/references/generated/imblearn.under_sampling.CondensedNearestNeighbour.html}} python library was used to perform CNN sampling. This resulted in a sampled data with 9907 sentences ($44\%$ of the original training data) as the high-quality training data. Since we retained all the positive data instances (check-worthy statements) during the data pruning process, the resulting high-quality training data comprised $54.6\%$ positive instances and $45.4\%$ negative instances.

\begin{figure}
  \centering
  \begin{subfigure}{0.48\textwidth}
    \centering
    \includegraphics[width=\linewidth]{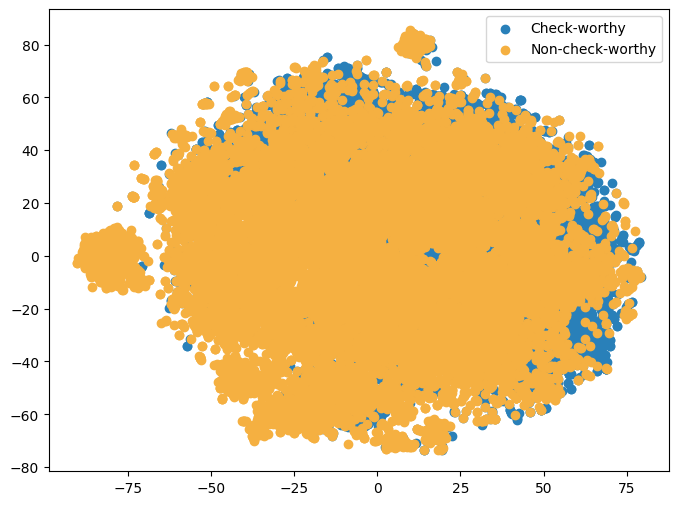}
    \caption{Original training data (before applying step 1 \& 2)}
    \label{fig:2d_all}
  \end{subfigure}
  \hspace{1cm}
  \begin{subfigure}{0.48\textwidth}
    \centering
    \includegraphics[width=\linewidth]{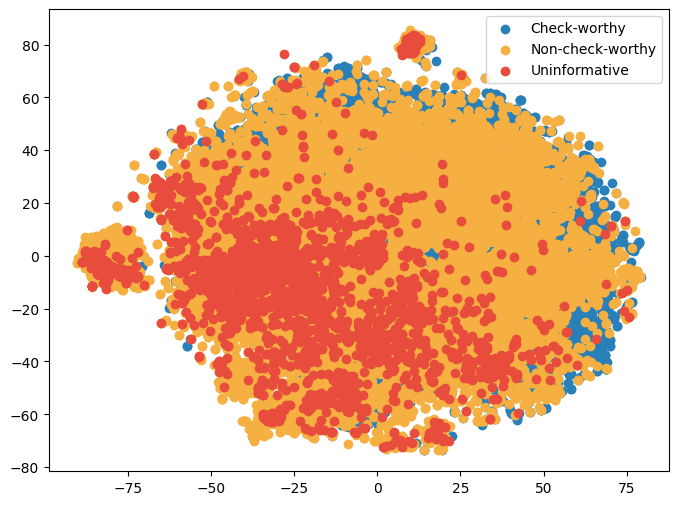}
    \caption{Original training data with uninformative sentences (filtered during step 1) highlighted}
    \label{fig:2d_informative}
  \end{subfigure}
  \begin{subfigure}{0.48\textwidth}
    \centering
    \includegraphics[width=\linewidth]{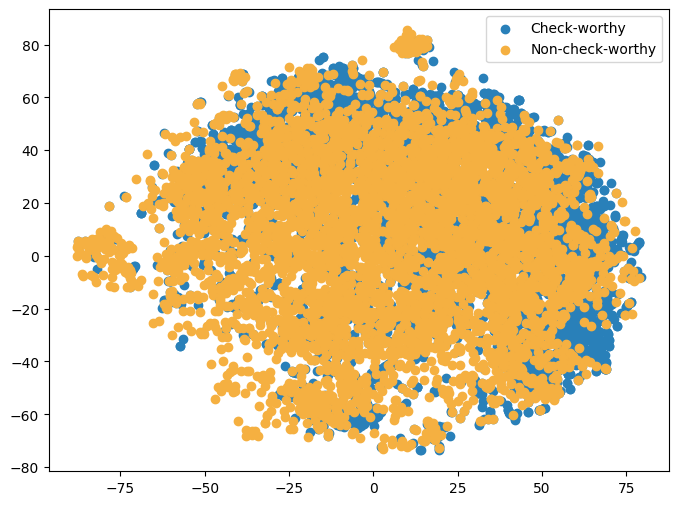}
    \caption{High-quality training data obtained after data pruning (after step 1 \& 2)}
    \label{fig:2d_cnn}
  \end{subfigure}
  \caption{2D visualization of the Training Data.}
  \label{fig:2d_visualization}
\end{figure}

Figure \ref{fig:2d_visualization} visualizes the training data points in 2 dimensions at each stage of the data pruning process. It can be observed that uninformative data points filtered during step 1 (Figure \ref{fig:2d_informative}) are concentrated around the bottom-left corner of the plot. Further, the CNN samples a similar distribution of non-check-worthy data points (Figure \ref{fig:2d_cnn}) for obtaining balanced training data.

\section{Results}\label{sec:Results}
We discuss the experiment results in this section, especially the hyper-parameters used to fine-tune the LLMs, the environment setting used, the performance of various LLMs with the consistency analysis, effect of prompt engineering, and the impact of training data pruning on check-worthy statement detection task.

\subsection{Hyper-parameters and Environment Setting}\label{sec:hyper-paramters}

\begin{table}[]
\centering
\caption{Hyper-parameters used for Fine-tuning.}
\label{tab:parameters}
\begin{tabular}{ll}
\toprule
Parameter                   & Value             \\ \midrule
Epochs                      & 3                 \\
Training batch size         & 2                 \\
Gradient accumulation steps & 2                 \\
Optimizer                   & Paged AdamW 32bit \\
Learning rate               & 2e-4              \\
Weight decay                & 0.001             \\
Maximum gradient norm       & 0.3               \\
Warmup ratio                & 0.03              \\
Temperature                 & 0.03              \\ \midrule
Lora Alpha                  & 16              \\
Lora dropout                  & 0.1              \\
Lora rank                  & 64              \\
\bottomrule
\end{tabular}
\end{table}

Hyper-parameters used to fine-tune the LLMs in our experiments are listed in Table \ref{tab:parameters}. Most of the hyper-parameters were the same for all the LLMs we experimented with, except the training batch size which was reduced to 1 for Mixtral, due to its memory demand. Consequently, the gradient accumulation steps for this model was increased to 4. All the experiments were conducted using Queen Mary's Apocrita HPC facility, supported by QMUL Research-IT \cite{king_2017_438045}. Specifically, 1 GPU (Volta V100 or Ampere A100) with 8 CPU cores, each composed of 11 GB memory was used to train and test all the models.

While generating the predictions by the fine-tuned models for evaluation, we noticed that a language model may generate different predictions for the same prompt and input sentence. Therefore, we ran each fine-tuned language model 5 times and obtained the majority prediction as the final prediction of the model. Further, we fine-tuned each model 3 times and reported the average performance of 3 fine-tuned models. We use the partitions \textit{train} and \textit{dev} for training and validation of the models, and the performance of the models is reported on the other two partitions \textit{dev-test} and \textit{test}. 

\subsection{Evaluation Metrics}
The official evaluation metric for CheckThat! 2024 task 1, check-worthiness estimation is F1-Score over the positive class. However, since one metric could have a bias toward the comparison, we report average accuracy, precision, and recall along with the F1-score. Further, we computed \textit{consistency@K} of the fine-tuned models indicating the fraction of data instances for which the model generated the same output class in all K iterations. \textit{Consistency} score report in the following subsections was computed over 5 iterations (consistency@5). 

\subsection{Comparison of LLMs}

\begin{table}[]
\centering
\caption{Performance of LLMs on the Test and Dev-Test Partitions.}
\label{tab:llm_comparison}
\begin{tabular}{lllllll}
\toprule
                    Partition      & Model              & Accuracy & Precision & Recall & F1-Score & Consistency \\
                          \midrule
\multirow{8}{*}{Test}     & Llama2-7b  & 0.905 $\pm$ 0.01    & 0.802 $\pm$ 0.021     & 0.841 $\pm$ 0.02  & \textbf{0.82} $\pm$ 0.019     & \textbf{0.936} $\pm$ 0.013       \\
                          & Llama2-13b & 0.897 $\pm$ 0.006   & 0.772 $\pm$ 0.021       & 0.856 $\pm$ 0.023   & 0.812 $\pm$ 0.009   & 0.921 $\pm$ 0.005      \\
                          & Llama3-8b  & \textbf{0.907} $\pm$ 0.002    & \textbf{0.863} $\pm$ 0.011    & 0.761 $\pm$ 0.011  & 0.809 $\pm$ 0.004     & 0.896 $\pm$ 0.002       \\
                          & Mistral       & 0.889 $\pm$ 0.005   & 0.747 $\pm$ 0.014     & \textbf{0.86} $\pm$ 0.017  & 0.799 $\pm$ 0.01    & 0.921 $\pm$ 0.01        \\
                          & Mixtral   & 0.891 $\pm$ 0.007  &  0.741 $\pm$  0.012  & 0.886 $\pm$ 0.011      &  0.807 $\pm$ 0.011     &  0.901 $\pm$ 0.011           \\
                          & Phi3-Mini-4K      & 0.897 $\pm$ 0  & 0.789 $\pm$ 0.007          & 0.822 $\pm$ 0.013        & 0.805 $\pm$ 0.003          & \textbf{0.936} $\pm$ 0.003            \\
                          & Falcon       & 0.891 $\pm$ 0         & 0.78 $\pm$ 0           &  0.806 $\pm$ 0      & 0.793 $\pm$ 0         & 0.912 $\pm$ 0 \\
                          & Gemma-7b & 0.9 $\pm$ 0.01 & 0.788 $\pm$ 0.014 & 0.841 $\pm$ 0.05 & 0.813 $\pm$ 0.018 & 0.909 $\pm$ 0.01 \\
                          \midrule
\multirow{7}{*}{Dev-Test} & Llama2-7b  & 0.944 $\pm$ 0.002    & 0.936 $\pm$ 0.005    & 0.898 $\pm$ 0.009 & 0.917 $\pm$ 0.003    & 0.95 $\pm$ 0.003      \\
                          & Llama2-13b & 0.941 $\pm$ 0.005    & 0.919 $\pm$ 0.013    & \textbf{0.907} $\pm$ 0.009  & 0.913 $\pm$ 0.007   & 0.923 $\pm$ 0.009       \\
                          & Llama3-8b  & 0.921 $\pm$ 0.003     & 0.956 $\pm$ 0.01     & 0.806 $\pm$ 0.009  & 0.874 $\pm$ 0.005    & 0.92 $\pm$ 0.002        \\
                          & Mistral       & 0.932 $\pm$ 0.016     & 0.909 $\pm$ 0.037     & 0.889 $\pm$ 0.018  & 0.899 $\pm$ 0.022    & 0.932 $\pm$ 0.001       \\
                          & Mixtral   & 0.939 $\pm$ 0.002 &   0.903 $\pm$ 0.004       & 0.92 $\pm$ 0.011  &  0.911 $\pm$ 0.003   & 0.93 $\pm$ 0.007     \\
                          & Phi3-Mini-4K       & \textbf{0.955} $\pm$ 0.004         & \textbf{0.961} $\pm$ 0.016           &  0.904 $\pm$ 0.005       & \textbf{0.932} $\pm$ 0.005         & \textbf{0.952} $\pm$ 0.007            \\
                          & Falcon       & 0.931 $\pm$ 0.0         & 0.967 $\pm$ 0          &  0.824 $\pm$ 0       & 0.89 $\pm$ 0         & 0.925 $\pm$ 0           \\
                          & Gemma-7b & 0.942 $\pm$ 0.002 & 0.942 $\pm$ 0.024 & 0.886 $\pm$ 0.021 & 0.913 $\pm$ 0.002 & 0.919 $\pm$ 0.024\\
                          \bottomrule
\end{tabular}
\end{table}

We compare the performance of the eight open-source LLMs in test and dev-test partitions and Table \ref{tab:llm_comparison} reports their average accuracy, precision, recall, F1-score, and consistency@5. Since the class distribution of dev-test and test partitions are different, we can observe that the performance in each partition varies for all the LLMs. While Phi3-Mini-4K obtains the highest F1-score in the dev-test portion, Llama2 models stand out as the best-performing models in the test partition based on F1-score. Further, both Llama2-7b and Phi3-Mini-4K demonstrate greater consistency in predicting class labels across both partitions compared to other models. On the other hand, Llama3-8b, one of the latest models from the Llama family reaches the highest accuracy and precision in the test partition. However, the model fails to outperform the other models in terms of F1-score due to poor recall. Similarly, Mixtral, the largest model we compared, fails to give a consistent performance across both partitions. Moreover, both Mistral and Falcon remain as the lower-performing models in both partitions.  

\begin{table}[]
\centering
\caption{Average Fine-tuning Time of LLMs.}
\label{tab:llm_running_time}
\begin{tabular}{ll}
\toprule
Model         & Average Fine-tuning Time \\
\midrule
Llama2-7b  & 6.3h                     \\
Llama2-13b & 11.9h                    \\
Llama3-8b  & 6.7h                    \\
Mistral       & 6.82h                    \\
Mixtral       & 33.86h                         \\
Phi3-Mini-4K    & 4h        \\
Falcon      & 6.26h \\
\bottomrule
\end{tabular}
\end{table}

Table \ref{tab:llm_running_time} lists the fine-tuning time of each LLM. Except for Mixtral and Phi3-Mini-4K, the largest and smallest models compared respectively, all the other models' fine-tuning time remains between 6-12 hours. As expected, the fine-tuning time increases with the number of parameters.

As we already mentioned, we were able to compare only the Llama models, Mistral, and Mixtral during the testing phase of the competition. Therefore, considering the performance of these models in terms of F1-score in the dev-test partition and the time and memory required to fine-tune the model, we chose Llama2-7b as the optimal LLM for the remaining experiments. 

\subsection{Consistency Analysis}
\begin{figure}
  \centering
  \includegraphics[width=0.75\linewidth]{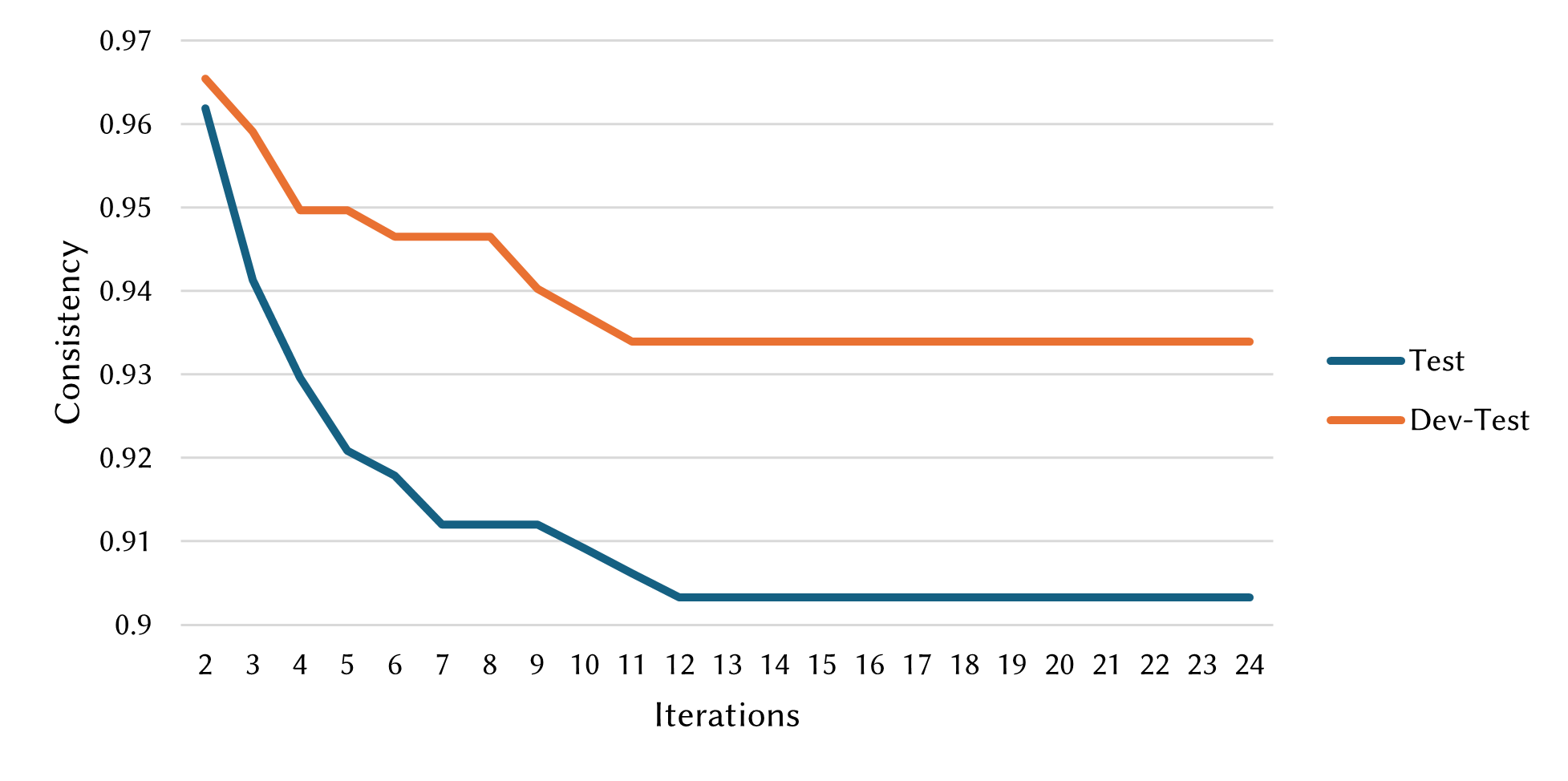}
  \caption{Consistency of Llama2-7b with Iterations}
  \label{fig:consistency}
\end{figure}

Since LLMs are text generation models, they tend to generate different output for the same input text even in a fine-tuned environment. Therefore, we analyze their consistency in predicting the same output class label over K iterations with the metric \textit{consistency@K}. Figure \ref{fig:consistency} presents the change in consistency with the number of iterations varied from 2 to 25 in the test and dev-test partition. It can be observed that the consistency declines with the increase in the number of iterations and becomes stable after around 11-12 iterations. While the model could reach a stable consistency in both partitions, the percentage of drop in consistency (difference between initial consistency when K=2, and the stable consistency) is nearly double for the test partition (3.1\% vs 5.9\%).     

\subsection{Effect of Prompt Engineering}

\begin{table}[]
\centering
\caption{Performance of Llama2-7b with Prompt Variation.}
\label{tab:performance_prompt}
\begin{tabular}{lllllll}
\toprule
                    Partition      & Prompt              & Accuracy & Precision & Recall & F1-Score & Consistency \\
                          \midrule
\multirow{4}{*}{Test}     & Prompt \ref{prompt-v1} & 0.905 $\pm$ 0.01    & 0.802 $\pm$ 0.021     & 0.841 $\pm$ 0.02  & 0.82 $\pm$ 0.019     & 0.936 $\pm$ 0.013      \\ 
& Prompt \ref{prompt-v2}  & \textbf{0.906} $\pm$ 0.003    & \textbf{0.807} $\pm$ 0.06     & 0.837 $\pm$ 0.007  & \textbf{0.823} $\pm$ 0.007     & 0.919 $\pm$ 0.003       \\
                          
                          & Prompt \ref{prompt-v3}  & 0.902 $\pm$ 0.002    & 0.782 $\pm$ 0.006    & \textbf{0.86} $\pm$ 0.013  & 0.82 $\pm$ 0.004     & \textbf{0.949} $\pm$ 0.003       \\
                          & No Instruction & 0.894  $\pm$ 0.012 & 0.801 $\pm$ 0.03 & 0.788 $\pm$ 0.017 & 0.794 $\pm$ 0.021 & 0.934 $\pm$ 0.018 \\
                          \midrule
\multirow{4}{*}{Dev-Test} & Prompt \ref{prompt-v1} & \textbf{0.944} $\pm$ 0.002    & \textbf{0.936} $\pm$ 0.005    & \textbf{0.898} $\pm$ 0.009 & \textbf{0.917} $\pm$ 0.003    & \textbf{0.95} $\pm$ 0.003      \\ 
& Prompt \ref{prompt-v2}  & 0.927 $\pm$ 0.006    & 0.892 $\pm$ 0.013    & 0.895 $\pm$ 0.005 & 0.894 $\pm$ 0.009    & 0.937 $\pm$ 0.003      \\
                          
                          & Prompt \ref{prompt-v3}  & 0.936 $\pm$ 0.008     & 0.91 $\pm$ 0.025     &  0.901 $\pm$ 0.005  & 0.906 $\pm$ 0.01    & 0.934 $\pm$ 0.005       \\
                          & No Instruction & 0.927 $\pm$ 0.008 & 0.926 $\pm$ 0.012 & 0.852 $\pm$ 0.016 & 0.887 $\pm$ 0.013 & 0.937 $\pm$ 0.003 \\
                          \bottomrule
\end{tabular}
\end{table}

We conducted experiments using three proposed versions of prompts discussed in Section \ref{sec:prompt-engeering} along with a prompt without any instruction to analyze the impact of prompt engineering on Llama2-7b model performance. Table \ref{tab:performance_prompt} presents the evaluation results on the test and dev-test partitions. While all three prompts proposed reach a similar F1-score in the test partition, their impact is quite evident in the dev-test partition. We can observe that Prompt \ref{prompt-v1} achieves the best overall scores across all metrics. It is worth noting, that the expanded prompt Prompt \ref{prompt-v3}, shows a notably high recall score and consistency in the test partition possibly due to the attention given to the pronouns in the instruction. Moreover, a substantial performance decline is observed when no instructions are included in the prompt highlights the importance of prompt engineering in achieving optimal results from LLMs. 

\begin{table}[]
\centering
\caption{Average fine-tuning time vs Instruction length in the Prompts.}
\label{tab:prompt_running_time}
\begin{tabular}{lll}
\toprule
Prompt         & Average Fine-tuning Time & Instruction Length (Number of Words) \\
\midrule
Prompt \ref{prompt-v1}  & 6.3h  & 60                    \\
Prompt \ref{prompt-v2} & 4.3h   & 11                  \\
Prompt \ref{prompt-v3}  & 7.1h  & 72                 \\
No Instruction & 3.1h & 0    \\
\bottomrule
\end{tabular}
\end{table}

In addition to performance metrics, Table \ref{tab:prompt_running_time} indicates that the fine-tuning time increases with the length of the instruction. According to the competition's settings, we primarily consider the F1-score performance of each prompt in the dev-test partition. Consequently, we selected Prompt \ref{prompt-v1} as the optimal one for the remaining experiments.

\subsection{Effect of Data Pruning}\label{sec:results_data_pruning}

\begin{figure}
  \centering
  \includegraphics[width=0.85\linewidth]{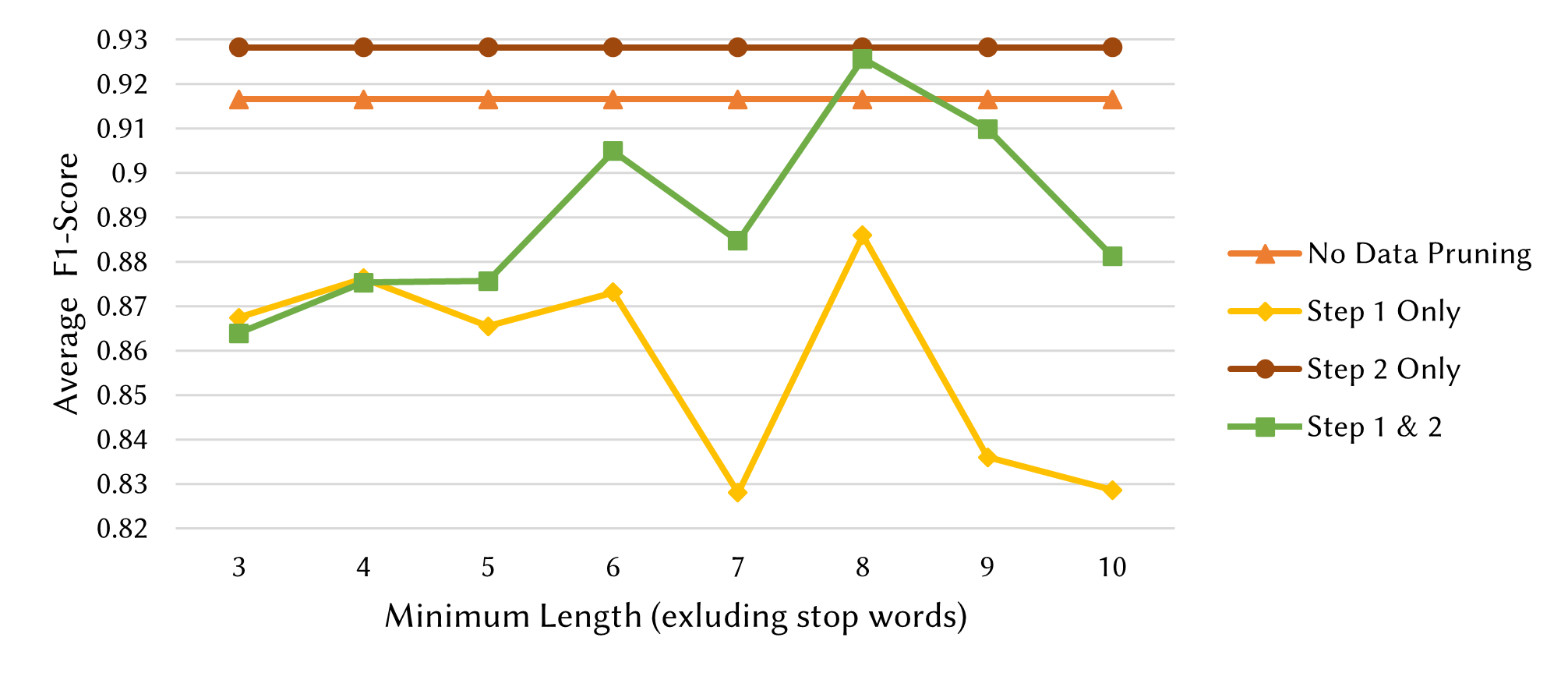}
  \caption{Average F1-score in Dev-Test partition with the change in \textit{minimum length} factor used for data pruning}
  \label{fig:f1_Score_variation}
\end{figure}

As we discussed earlier, we leave the \textit{minimum length} factor as a parameter of the data pruning process. Therefore we varied the \textit{minimum length} from 3-10 and observed the performance of the Llama2-7b model fine-tuned on the pruned dataset. Figure \ref{fig:f1_Score_variation} shows the F1-score of the fine-tuned model in the dev-test partition. It can be observed that Step 1 alone is not sufficient enough to identify high-quality training data, and Step 2 always boosts the performance when combined with Step 1 except when the \textit{minimum length} factor is set to very low (3-4). The optimal performance for the two-step data pruning is obtained when the \textit{minimum length} factor is set to 8. 

\begin{table}[]
\centering
\caption{Performance of Llama2-7b with Data Pruning.}
\label{tab:performance_pruning}
\begin{tabular}{lllllll}
\toprule
                    Partition      & \begin{tabular}[c]{@{}l@{}}Pruning\\ Technique\end{tabular}               & Accuracy & Precision & Recall & F1-Score & Consistency \\
                          \midrule
\multirow{4}{*}{Test}     & None & 0.905 $\pm$ 0.01    & 0.802 $\pm$ 0.021     & 0.841 $\pm$ 0.02  & \textbf{0.82} $\pm$ 0.019     & 0.936 $\pm$ 0.013       \\ 
& Step 1 only & 0.886 $\pm$ 0.005 & 0.802 $\pm$ 0.004 & 0.739 $\pm$ 0.02 & 0.769 $\pm$ 0.013 & \textbf{0.943} $\pm$ 0.004         \\
& Step 2 only & \textbf{0.907} $\pm$ 0.004  & \textbf{0.831} $\pm$ 0.007     & 0.803 $\pm$ 0.013     & 0.817 $\pm$ 0.01  & 0.924 $\pm$ 0.005  \\
& Step 1 \& 2 & 0.891 $\pm$ 0.008     & 0.752 $\pm$ 0.016      & \textbf{0.86} $\pm$ 0.013   & 0.802 $\pm$ 0.015 & 0.904 $\pm$ 0.003         \\
                          \midrule
\multirow{4}{*}{Dev-Test} & None & 0.944 $\pm$ 0.002    & 0.936 $\pm$ 0.005    & 0.898 $\pm$ 0.009 & 0.917 $\pm$ 0.003    & \textbf{0.95} $\pm$ 0.003      \\ 
                          
& Step 1 only & 0.929 $\pm$ 0.016 & \textbf{0.967} $\pm$ 0.012 & 0.818 $\pm$ 0.042 & 0.886 $\pm$ 0.027 & 0.945 $\pm$ 0.012       \\
& Step 2 only & \textbf{0.953} $\pm$ 0.003    & 0.96 $\pm$ 0.009      &  0.898 $\pm$ 0.009  & \textbf{0.928} $\pm$ 0.005 & 0.943 $\pm$ 0.003       \\
& Step 1 \& 2 & 0.95 $\pm$ 0.005  & 0.929 $\pm$ 0.01     & \textbf{0.923} $\pm$ 0.005      & 0.926 $\pm$ 0.008   &  0.928 $\pm$ 0.003       \\
                          \bottomrule
\end{tabular}
\end{table}

Table \ref{tab:performance_pruning} presents the performance of Llama2-7b on the original and pruned training data. It can be observed that the model trained using only the step 2 data pruning approach yields the highest F1-score and accuracy in the dev-test partition and its performance in the test partition is close to the model trained without any data pruned. While the two-step data pruning approach reaches a slightly lower F1 score in the test partition, it stands out as the high recall model in both partitions. Further, it is worth noting that this model resulted in a better precision-recall trade-off in the dev-test partition compared to the models trained with individual pruning steps (step 1 only and step 2 only). Moreover, the consistency of the models in predicting the same output class ranged from 0.9-0.95, while the lower range is always observed in the dev-test partition. During the testing phase of the completion, we could not compare the average performance of the models due to time limitations. Therefore, we submitted the class labels predicted by the Llama2-7b model fine-tuned without the data pruning process for the CheckThat! 2024 task1 leader board, as it yielded a slightly higher F1-score compared to other approaches. This submission was ranked 1st place in the leaderboard with the highest F1-score of 0.802. While this score is lower than the average F1-score reported in Table \ref{tab:performance_pruning}, the standard deviation indicates that the model performance could vary from 0.801 to 0.839.

\begin{table}[]
\centering
\caption{Training Data Size and Average Fine-tuning Time with Data Pruning.}
\label{tab:datapruning_running_time}
\begin{tabular}{llll}
\toprule
Pruning Technique         & Training data size & \begin{tabular}[c]{@{}l@{}}\% of data retained\\during pruning\end{tabular} & Average Fine-tuning Time \\
\midrule
None  & 22,499 &    100\% & 6.3h                     \\
Step 1 only & 20,141 &  89\%    & 5.7h  \\
Step 2 only & 10,227 &  45.5\%    & 3h \\
Step 1 \& 2 & 9,907 &  44\%     & 2.9h \\
\bottomrule
\end{tabular}
\end{table}

Table \ref{tab:datapruning_running_time} reports the training data size and their corresponding average fine-tuning time. This demonstrates that using data pruning strategies allows competitive performance on both test and dev-test data while utilizing only about 44\%-44.5\% of the original training data. Further, the fine-tuning time is reduced in a similar proportion, cutting the training time by more than half compared to using the original training data. This indicates that obtaining high-quality training data is crucial for developing effective check-worthy statement detection models from political transcriptions.

\section{Conclusion}\label{sec:conclusion}
This paper demonstrates the experiments conducted by the FactFinders team for CheckThat! 2024 task 1, check-worthiness estimation in English. We experimented with eight open-source LLMs with fine-tuning and prompt engineering to identify check-worthy statement detection from political transcriptions. Results using our Llama2-7b model fine-tuned on the training data secured the 1st position in the leaderboard among a total of 26 participants, with F1-scores surpassing the baseline for the task. This demonstrates that open-source models are powerful in check-worthy statement detection in the English language. Further, we demonstrated the role of data pruning in identifying high-quality training data for effective learning. Our results show that competitive or better performance can be obtained by utilizing only about 44\% of training data by saving the fine-tuning time in a similar proportion. Apart from the fine-tuned LLMs for check-worthy statement detection, we utilized LLMs for refining prompts and identifying informative verbs in zero-shot setting.

The key challenges we faced while utilizing LLMs for check-worthy statement detection were the memory requirements and their inconsistent response in predicting the class label for the same input statement. We used the Low-Rank Adaption technique (LoRA) for effective fine-tuning with low GPU memory usage to conquer the memory requirements. While we tried to overcome the consistency issue by running the fine-tuned models 5 times and obtaining the majority prediction, our consistency analysis reveals that the consistency score itself is unstable during early iterations, and may have a drop of around 6\% while reaching stability. This behavior of LLMs highly questions their adaptation for classification tasks in general as inconsistent responses may result in less reproducible results. 

\section*{Acknowledgments}
Rrubaa Panchendrarajan is funded by the European Union and UK Research and Innovation under Grant No. 101073351 as part of Marie Skłodowska-Curie Actions (MSCA Hybrid Intelligence to monitor, promote, and analyze transformations in good democracy practices). Yufeng Li is funded by China Scholarship Council (CSC).

\bibliography{references}

\end{document}